\documentclass[10 pt, conference]{ieeeconf}  

\IEEEoverridecommandlockouts                              

\usepackage{amssymb,url,amsmath,bm}
\usepackage{algorithm}
\usepackage{graphicx}
\usepackage{epstopdf}
\usepackage{textcomp}
\usepackage{color}
\usepackage{ifpdf}
\usepackage{url}
\usepackage{enumerate}
\usepackage{array}
\usepackage{flushend}
\usepackage{multirow,booktabs}
\usepackage[pdfstartview=]{hyperref}
\usepackage[export]{adjustbox} 
\usepackage{subfig}
\usepackage{mdwlist}
\usepackage[utf8]{inputenc}
\usepackage{todonotes}

\usepackage{tcolorbox}
\newtcolorbox[auto counter]{algorithmbox}[2][]{colback=red!5!white,colframe=red!75!black,fonttitle=\bfseries, title=\alg\thetcbcounter: #2,#1}

\graphicspath{{./media/img/}}

\newcommand{\alg}[1]{Algorithm~\ref{#1}}

\newcommand{\eq}[1]{Eq.~\ref{#1}}
\newcommand{\fig}[1]{Fig.~\ref{#1}}

\newcommand{\sect}[1]{Sec.~\ref{#1}}

\DeclareMathOperator*{\argmin}{arg\,min}

\usepackage{cite}

\title{\LARGE \bf
2D Linear Time-Variant Controller for Human's Intention Detection for Reach-to-Grasp Trajectories in Novel Scenes}

\author{Claudio Zito$^{1,2,\ast}$, Tomasz Deregowski$^{1,\ast}$ and Rustam Stolkin$^{2}$
\thanks{This work was supported by UK Engineering and Physical Sciences Research Council (EPSRC No. EP/R02572X/1) for the National Centre for Nuclear Robotics (NCNR).}
\thanks{$^{\ast}$ First co-author}%
\thanks{$^{1}$ IRLab, School of Computer Science, University of Birmingham, UK}%
 \thanks{$^{2}$ ERL, School of Metallurgy and Materials, University of Birmingham, UK,
 {\tt\small C.Zito@bham.ac.uk}}
}

\begin{document}

\maketitle
\thispagestyle{empty}
\pagestyle{empty}

\begin{abstract}

Designing robotic assistance devices for manipulation tasks is
challenging. This work is concerned with improving accuracy and usability of semi-autonomous robots, such as human operated manipulators or exoskeletons. The key insight is to develop a system that takes into account context- and user-awareness to take better decisions in how to assist the user. The context-awareness is implemented by enabling the system to automatically generate a set of candidate grasps and reach-to-grasp trajectories in novel, cluttered scenes. The user-awareness is implemented as a linear time-variant feedback controller to facilitate the motion towards the most promising grasp. Our approach is demonstrated in a simple 2D example in which participants are asked to grasp a specific object in a clutter scene. Our approach also reduce the number of controllable dimensions for the user by providing only control on $x-$ and $y-$axis, while orientation of the end-effector and the pose of its fingers are inferred by the system. The experimental results show the benefits of our approach in terms of accuracy and execution time with respect to a pure manual control.   

\end{abstract}


\section{INTRODUCTION}\label{sec:introduction}
Semi-autonomous robots operated by human subjects to augment, substitute or assist motor functions are rapidly gaining ground becoming a mainstream type of technology. Their application is extensive in several domains: robots operate semi-autonomously in hazardous environments while the user is in safety, e.g. nuclear waste disposal~\cite{bib:fulbright_1995}, but also exoskeletons and prostheses are used by patients with neuromuscular disorders to regain motor functions~\cite{bib:cordella_2016, bib:gopura_2009}. Though, current human-machine interfaces (HMIs) are still not capable enough to control these devices effectively and to their full potential. So it results in low take-up, high cognitive burden for the users, poor coordination, long training session and slow operation speeds. This is particularly noticeable in upper-limb robots, due to the inherent complexity and ample possibilities of arm and hand movements~\cite{bib:cordella_2016}. 

Our work is a preliminary investigation towards context- and user-awareness in semi-autonomous manipulators. We are concerned with how to reduce the user's burden and the training session time for a semi-autonomous robot manipulator by making the system more intuitive for the user. In contrast to previous work, e.g.~\cite{bib:kent_2017}, this is done by adding an active AI component in the semi-autonomous system instead of solely improving the user's feedback, e.g.~\cite{zito_2013,zito_w2012,zito_w2013,rosales_2018}.  

\begin{figure}[!t]
\centerline{
\subfloat[]{\includegraphics[width=.19 \textwidth]{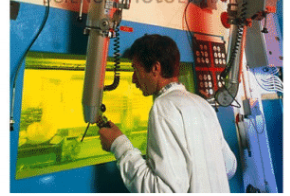}\label{fig:operator}}
\subfloat[]{\includegraphics[width=.29 \textwidth,height=.13\textwidth]{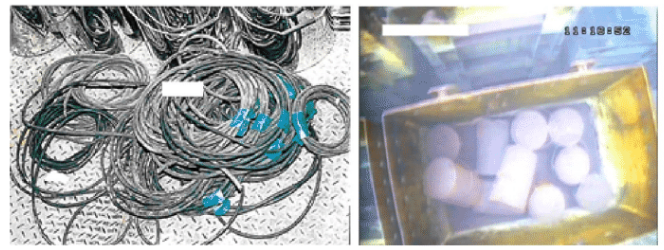}\label{fig:waste}}
}
\caption{The figure shows: (a) current workspace of the human operator for the sort and segregate application. The operator has to look through a small window to operate the mechanical arms; and (b) typical waste items for a nuclear facility: wiry, free-form and deformable objects. Pictures courtesy of UK National Nuclear Lab Workington test facility. Best seen in colours.
}
\label{fig:topology}
\end{figure}

The key insight of this paper is that by making the system user- and context-aware we will provide better usability and accuracy. This work is mainly focused on grasping in cluttered scenes. We improve context-awareness by employing a generative model similar to~\cite{kopicki2015oneshot,zito_2019,stuber_2019} to estimate a set of possible candidate grasps in the scene, and for each candidate grasp we plan a reach-to-grasp trajectory. The user's motion commands are then used to detect his/hers intention and facilitate the motion towards the selected target. To do so, we proposed a linear time-variant LQR controller that filters the motion command along the candidate trajectory. In other words, the LTV-LQR creates a basin of attraction towards the most promising trajectory. In addition, our system provides the user with controllability over 2D ($x-$ and $y-$axis) while the orientation of the end-effector and the pose of the fingers is automatically inferred from the selected trajectory. We also encode in the LTV-LQR a recovering method to identify when the user to move away from an undesired trajectory and guide him/her towards a new best candidate. 

Although several other researchers have investigated LQR approaches for stabilising human-like motions, e.g.~\cite{Alvarez2017, todorov2003}, to our knowledge this is the first time that such an approach is used to identify human's intention and to improve reach-to-grasp actions.  




\section{BACKGROUND}\label{sec:background}
In this section we present the basic components that we use for generating grasps and approaching trajectories on a clutter scene. First, we introduce the concept of 2D surface features, which are used to learn a contact. Contacts are then encoded as kernels to build a full joint probability density (PDF) of robot's links and of surface features. While this idea was firstly introduced by us in~\cite{kopicki2015oneshot} to learn 3D contacts, in this work we present a simplified version in 2D. The contacts are a generative models that allows us to generate candidate grasps on a clutter scene. Finally we introduce the concept of an linear time-variant LQR controller that is used to create manifolds along the trajectories. Such manifolds enable us to detect the user's intention and guide him/her towards the selected target. 

\subsection{Surface features}\label{sec:surface_features}

To learn a probabilistic representation of a contact we rely on local object surface features. Features are composed of a  2D position, a 1D orientation, and a 1D local surface descriptor that encodes the local curvature. A feature belongs to the space $SE(2) \times \mathbb{R}$ , where $SE(2)$ is the special Euclidian space that describes a rigid body pose by $x-$, $y-$axis and a rotation over the $z-$axis, $\theta$. The surface descriptors are composed of a scalar. We thus represent an object as a set $S$ of $N_{S}$ surface features $s_j$
\begin{equation}
S = \{s_j : s_j \in SE(2) \times \mathbb{R}\}_{j\in[1,N_{S}]}.
\end{equation}
Let us denote the separation of a feature $s$ into $\mathbf{p}\in\mathbb{R}^2$ for position, a $\theta $ is an angle in radian for orientation, and $r \in \mathbb{R}$ for the surface descriptor.  For compactness, we also denote the pose of a feature as $v$. As a result, we have $s = (v, r)$, and $v = (\mathbf{p}, \theta)\in SE(2)$.

Surface descriptors correspond to the local principal curvature \cite{spivak1999geometry}. The surface normal at $\mathbf{p}$ is computed from the nearest neighbours of $\mathbf{p}$ using a PCA-based method, e.g. \cite{kanatani2005geometry}.  The curvature at point $\mathbf{p}$ is encoded along the plane tangential to the object's surface, $r$, i.e. perpendicular to the surface normal at $\mathbf{p}$. The surface normal and principal direction allow us to define a 2D reference frame that is associated to a point $\mathbf{p}$. 

\subsection{Kernel density estimation}

In this paper, we represent PDFs non-parametrically with a set of features, or particles. The underlying PDF is created through \textit{kernel density estimation} \cite{silverman1986density}, by assigning a kernel function to each particle supporting the density. For contact models, we consider PDFs defined on surface features $s$ (Sec.~\ref{sec:surface_features}).  For that purpose, let us denote by $\bm{\mu}$ a surface feature vector given by $\bm{\mu}  = (\bm{\mu}_p , \mu_ \theta , \mu_r )$, and by $\bm{\sigma}$ a vector of real numbers $\bm{\sigma} = (\bm{\sigma}_p , \sigma_ \theta , \sigma_r )$. Where $\bm{\mu}_p=(\mu_x,\mu_y)$ is the position mean, and $\bm{\sigma}_p=(\sigma_x,\sigma_y)$ is the diagonal of the covariance. We thus define our kernel as
\begin{equation}
\mathcal{K}(s \mid \bm{\mu}, \bm{\sigma}) = \mathcal{N}_ 2 (\mathbf{p}\mid\bm{\mu_p} , \bm{\sigma}_p )\mathcal{N}_1(\theta\mid\mu_ \theta , \sigma_ \theta )\mathcal{N}_1 (r\mid\mu_r , \sigma_r ),
\end{equation}
where $\mu$ is the kernel mean point, $\sigma$ is the kernel bandwidth, $\mathcal{N}_n$ is an $n$-variate isotropic Gaussian kernel. Given a set of $K$ surface features, the probability density in a region of space is then determined by the local density of the particles in that region, as
\begin{equation}
P(s) \simeq \sum_{j=i}^{K} w_j \mathcal{K}(s \mid s_j , \bm{\sigma}),
\end{equation}
where $\bm{\sigma}\in\mathbb{R}^4$ is the kernel bandwidth and $w_j \in \mathbb{R}^+$ is a weight associated to $s_j$ such that $\sum_{j}w_{j} = 1$. 

\subsection{Linear Time-Variant LQR}

Linear time-variant LQR (LTV-LQR) is a mathematically elegant solution for controlling non-linear dynamical system~\cite{Bertsekas2000}. The underlying idea is to locally linearise the dynamics of the system around a feasible trajectory. By feasible we mean that the trajectory is a solution of the dynamics. 
Let us consider the non-linear dynamic system
\begin{equation}\label{eq:system}
\mathbf{\dot{x}}=f(\mathbf{x},\mathbf{u})
\end{equation}
and its Taylor expansion around a point $(\mathbf{x_0},\mathbf{u_0})$, which results in
\begin{equation}\label{eq:taylor}
\begin{split}
\dot{\mathbf{x}}\approx &f(\mathbf{x_0},\mathbf{u_0})+\frac{\partial f}{\partial\mathbf{x}}(\mathbf{x}-\mathbf{x_0})+\frac{\partial f}{\partial\mathbf{u}}(\mathbf{u}-\mathbf{u_0})\\
&=\mathbf{c}+\mathbf{A}(\mathbf{x}-\mathbf{x_0})+\mathbf{B}(\mathbf{u}-\mathbf{u_0})
\end{split}
\end{equation}
Following the standard notation, $\mathbf{x}\in\mathbb{R}^n$ represents a state, $\mathbf{u}\in\mathbb{R}^m$ an input control, $\mathbf{\dot{x}}\in\mathbb{R}^n$ is the state change, $\mathbf{A}\in\mathbb{R}^{n\times n}$ and $\mathbf{B}\in\mathbb{R}^{n\times m}$ describe the transition dynamics of the system and how the effects of the input control over states, respectively.

By parametrising the system over time we change coordinates to
\begin{equation}\label{eq:change_coordinates}
\begin{split}
\mathbf{\bar{x}}(t)&=\mathbf{x}(t)-\mathbf{x_0}(t)\\
\mathbf{\bar{u}}(t)&=\mathbf{u}(t)-\mathbf{u_0}(t)
\end{split}
\end{equation}
in which $[\mathbf{x_0}(t),\mathbf{u_0}(t)]$ is a feasible trajectory. We can then rewrite the system as 
\begin{equation}\label{eq:dynamics}
\mathbf{\dot{\bar{x}}}(t)=\mathbf{\dot{x}}(t)-\mathbf{\dot{x}_0}(t)=\mathbf{\dot{x}}(t)-f(\mathbf{x_0}(t),\mathbf{u_0}(t))
\end{equation}
and by using \eq{eq:taylor}
\begin{equation}\label{eq:tv_lqr}
\begin{split}
\mathbf{\dot{\bar{x}}}(t)=&\frac{\partial f_0}{\partial\mathbf{x}}(\mathbf{x}(t)-\mathbf{x_0}(t))+\frac{\partial f_0}{\partial\mathbf{u}}(\mathbf{u}(t)-\mathbf{u_0}(t))\\
=&A(t)\mathbf{\bar{x}}(t)+B(t)\mathbf{\bar{u}}(t)
\end{split}
\end{equation} 
such that the linearisation of the dynamics, i.e. the reference of the coordinate system, moves along the trajectory. In \eq{eq:tv_lqr} we used a compacted notation $f_0$, instead of $f(\mathbf{x_0}(t),\mathbf{u_0}(t))$, to write the partial derivatives of $f$ at point $(\mathbf{x_0}(t),\mathbf{u_0}(t))$.

The trajectory stabilisation is formulated as minimising the following finite horizon cost function in quadratic form,
\begin{equation}\label{eq:quadratic}
\begin{split}
J(\mathbf{\bar{x}},0)=&\frac{1}{2}\mathbf{\bar{x}}(0)^\top Q(0)\mathbf{\bar{x}}(0)\\
&+\frac{1}{2}\sum_{t=N}^{1}(\mathbf{\bar{x}}(t)^\top Q(t)\mathbf{\bar{x}}(t)+\mathbf{\bar{u}}(t)^\top R(t)\mathbf{\bar{u}}(t))
\end{split}
\end{equation}
where the top line encodes a penalising factor for not reaching the goal state, $\mathbf{x}(0)$, at the end of the trajectory. The bottom line penalises the system at time $t$ for being away from $\mathbf{x_0}(t)$. The state cost-weighting matrices $Q(t)$, for $t\geq0$, need to be symmetric positive semi-definite, while the control cost-weighting matrix $R(t)$ need to be positive definite. Our implementation of the state and control cost-weighting matrices is shown in \sect{sec:ltv_lqr}.

The feedback control matrix, $K(t)\in\mathbb{R}^{m\times n}$, is used to compute the optimal control input at time $t$ that minimises the cost function in~\eq{eq:quadratic}, such that
\begin{equation}\label{eq:feedback}
\mathbf{u}^*(t)=\mathbf{u_0}(t)-K(t)\mathbf{x}(t)
\end{equation}
and K is computed as
\begin{equation}\label{eq:K}
K_t=R(t)^{-1}B(t)^\top P(t)
\end{equation} 
$P(t)$ is found by solving the Riccati Differential Equation~\cite{Bertsekas2000}.



\section{OUR APPROACH}\label{sec:approach}
In this section we describe our approach. Given a novel scene, we generate a set of possible candidate grasps and, for each grasp, a reach-to-grasp trajectory for the gripper. We then compute a sequential feedback controller for each trajectory. When the user's input comes in, the system selects the best feedback controller that matches the input and assist the user along the selected trajectory. If the user moves away from the selected trajectory, the effects of the feedback controller gradually disappear until a new feedback controller will be selected.

\subsection{Contact model}\label{sec:contact_model}

\begin{figure}[!t]
\centerline{
\subfloat[]{\includegraphics[width=.245 \textwidth]{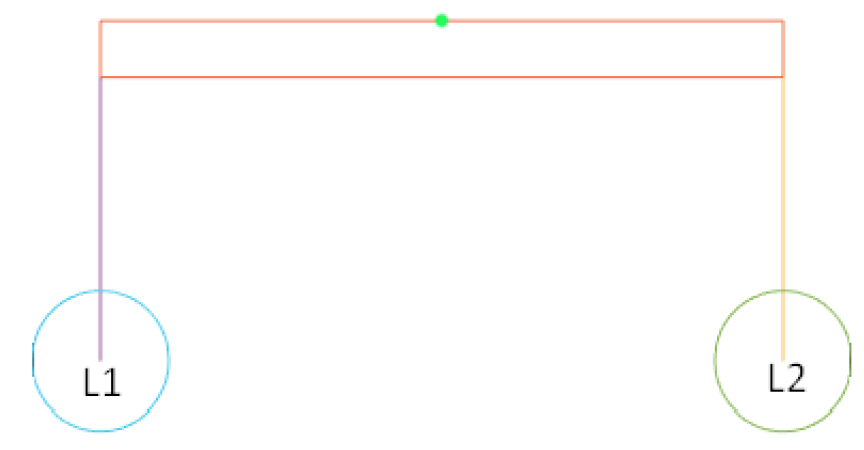}\label{fig:gripper}}
\subfloat[]{\includegraphics[width=.245 \textwidth]{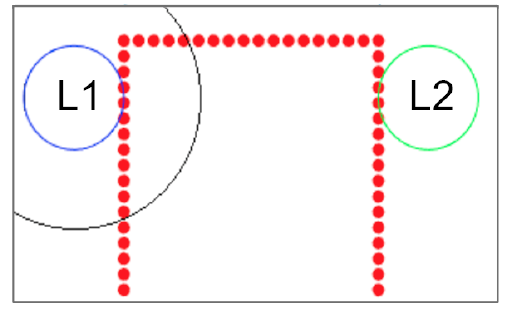}\label{fig:contact_model}}
}
\caption{(a) A simulated two spherical fingers parallel gripper. The gripper is composed by one link for each finger, L1 and L2. (b) The learnt contact model as a pinch grasp on a rectangular shaped object. The black circle around link L1 represents the neighbour in which surface's features are considered for learning the contact. Best seen in colours.
}
\label{fig:contact}
\end{figure}

To generate a set of possible candidate grasp we first need to learn a contact model. A contact model is a full joint probability of the pose of the gripper's links and of an object's surface features. It describes the relations between the gripper's links and the object's surface involved in the contact. Figure~\ref{fig:contact} shows a graphical representation of (a) the gripper and (b) a pinch grasp contact model.

\begin{figure}[!t]
	\centering
	\subfloat[]{\includegraphics[width=.4 \textwidth]{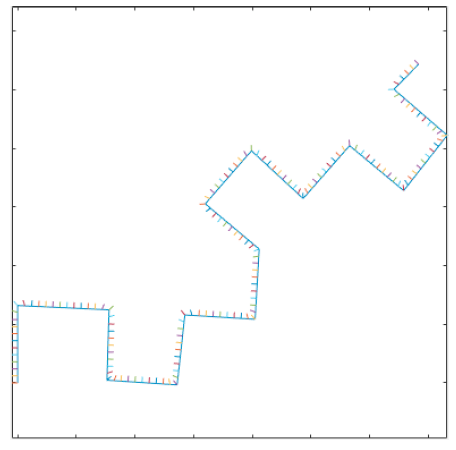}\label{fig:topology}}\\
	\subfloat[]{\includegraphics[width=.4 \textwidth]{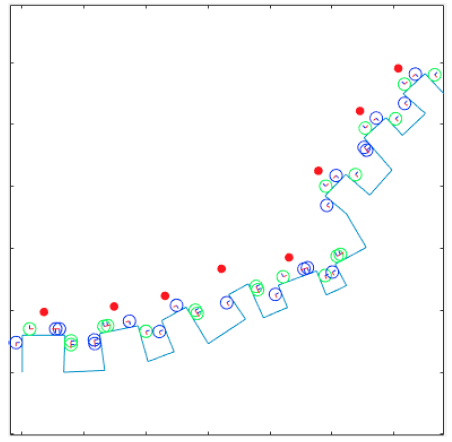}\label{fig:grasps}}
	
	\caption{Our simulated 2D scene. Figure (a) shows the landscape of a possible scene with surface's normals computed. Figure (b) shows a set of link's poses, blue circles for L1 and green for L2, generated by the query density. The red points identify the candidate grasps, as a 2D pose for the reference frame of the gripper, sampled from the query density which are configurations of the gripper where both links generate a contact. Best seen in colours.
	}
	\label{fig:simulation}
\end{figure}

Given a 2D description of the clutter scene $O$ as a set of $N_O$ surface features $\{s_j\}_{j=1}^{N_O}$ (\sect{sec:surface_features}), a contact model $C^{L_i}$ is constructed for the $i^{th}$ link. Surface features close to the link surface are considered more important than those lying far from the surface, thus the features are weighted, to make their influence on $C^{L_i}$ decrease exponentially with the square distance to the link.  Let us denote by $z_{j} = (\mathbf{p}_{j}, \theta_{j})$ the pose of $L$ relative to the pose $v_j$ of the $j$\textsuperscript{th} surface feature. In other words, $z_{j}$ is defined as
\begin{equation}
z_{j} = v_j^{-1} \circ l,
\end{equation} 
where $l$ denotes the pose of $L_i$ in the world frame, $\circ$ denotes the pose composition operator, and $v_j^{-1}$ is the inverse of $v_j$. The contact model is then estimated as

\begin{align}
C^{L}&(l, r) \simeq \nonumber \\
&\frac{1}{Z} \sum_{j=1}^{N_{O}} w_{j}\mathcal{N}_2(\mathbf{p}\mid \bm{p}_{j}, \bm{\sigma}_p)\mathcal{N}_1(\theta \mid \theta_{j}, \sigma_ \theta)\mathcal{N}_1(r \mid r_j, \sigma_r),
\end{align}where $Z$ is a normalising constant, and $z = (\mathbf{p}, \theta)$. 

\subsection{Query density}\label{sec:query}
	
A query density results from the combination of a contact model with a novel scene $O$. The scene in this work is represented by a 2D landscape, see \fig{fig:topology}. The purpose of a query density is both to generate and evaluate poses of the corresponding gripper's link on the new scene (\fig{fig:grasps}).

A query density $M^{L_i}$ is a PDF defined as   
\begin{equation}
M^{L_i}=P(l, z, v, r),
\end{equation}where $v$ denotes a point on a object's surface in the scene, expressed in the world frame, $r$ is the surface curvature of such a point, $z$ denotes the pose of a link relative to a local frame on the object, and $l$ is the absolute pose in the world frame of the gripper's link, $L_i$.

At prediction time, we use a query density $M^{L_i}$ to generate poses of the gripper's link on the new scene. We achieve this by marginalising with respect to $z$, $v$, and $r$, obtaining the distribution $M^{L_i}(l)$ which models the pose $l$ in the world frame of the link $L_i$.  We approximate the query density by $K_{M^{L_i}}$ kernels centred on the set of weighted robot link poses. As proposed in \cite{kopicki2015oneshot}, by sampling from $M^{L_i}$ for each link, a grasp distribution is obtained from which is possible to compute a set of candidate grasps for the new scene. The same learnt contact model allows us to generalise similar surfaces in the novel scene and to adapt the grasp to different sizes of the objects, \fig{fig:grasps}.

\subsection{LTV-LQR controller}\label{sec:ltv_lqr}

\begin{figure}[t]
\centerline{
\includegraphics[width=.3 \textwidth]{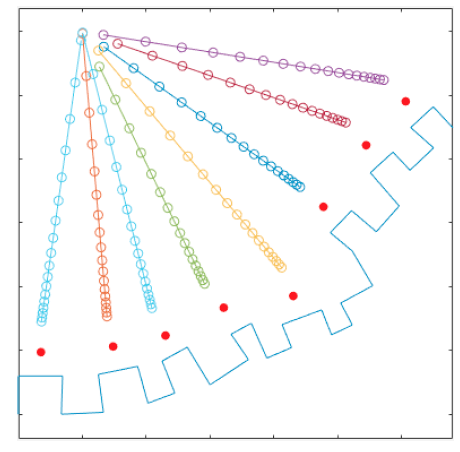}
}
\caption{The red point in thew figure represents the candidate grasps, as a 2D pose for the gripper. The system computes a straight line trajectory for each grasp. The trajectory is discretised in a set of waypoints, $\mathbf{x_g}(t)$, in which the local feedback controller in \eq{eq:tv_lqr} is centred. Best seen in colours.
}
\label{fig:trajectories}
\end{figure}

For each grasp, $g$, computed by the query density, as described in \sect{sec:query}, we compute a reach-to-grasp trajectory, $[\mathbf{x_g}(t),\mathbf{u_g}(t)]$, from the current pose of the gripper, $\mathbf{x_g}(0)$. Note that our approach is independent from the method used to generate the trajectory. In our demonstration the trajectories are computed as straight lines to the goal, as shown in \fig{fig:trajectories}. We assumed no obstacles in the free space. However the orientation and aperture of the gripper is computed along the trajectories so that no collision will happen with the skyline of the scene. During a grasp, the gripper closes to generate contacts but penetration is not allowed.       

First, the state space in our approach is defined in terms of 2D position and 1D orientation, i.e. $SE(2)$. Without losing generality, we describe a state $\mathbf{x}=[x,y, \theta]^\top$ as a three-element vector, and the control input in terms of velocity, $\mathbf{u}=[\dot{x},\dot{y}]$. The end-effector orientation, $\theta$, and the pose of the fingers is controlled directly by the system by super-imposing the orientation and the finger's pose planned in the selected trajectory. Using the change in coordinates defined in \eq{eq:change_coordinates} we obtain the system formulation in \eq{eq:tv_lqr}.

The cost-weighted matrices $Q(t)$ and $R(t)$ are designed to drive the user towards the selected trajectory, however their influence need to disappear when the system realises that the user's intention is to move towards another candidate grasp. This is done by an exponential function as follows 
\begin{align}
Q(t)=&
\begin{bmatrix}
e^{-t} & 0 & 0 \\
0 & e^{-t} & 0\\ 
0 & 0 & 1 
\end{bmatrix}\\
R(t)=&
\begin{bmatrix}
e^{-t} & 0\\
0 & e^{-t} &
\end{bmatrix}
\end{align} 
where $Q(0)$ is the cost-weighted matrix associated with the final step in \eq{eq:quadratic}.   

Finally we define how the user's input is integrated in \eq{eq:feedback}. Typically LTV-LQR computes the difference between the ideal control, $\mathbf{u_0}$, at time $t$ and the feedback computed by the matrix $K$ with respect to the state, $x$, at time $t$. We extend this formulation to make a prediction about where the user wants to go and how to cope with that. First, the prediction is done by integrating the user's input, $\mathbf{u}$, in the state space by
\begin{equation}\label{eq:prediction}
\phi(\mathbf{x},\mathbf{u},t)=\int_{t}^{t+1}f(\mathbf{x}(t),\mathbf{u}(t))dt
\end{equation} 
and then we select the feedback controller with the lowest immediate cost function for the predicted state at time $t$. Let $\mathbf{\hat{x}}=\phi(\mathbf{x},\mathbf{u},t)-\mathbf{x_g}(t)$ be the difference between the predicted state and the desired state for the trajectory associated to the grasp $g$. Similarly, let $\mathbf{\hat{u}}=\mathbf{u}(t)-\mathbf{u_g}(t)$ be the difference between the user input and the expected input for the trajectory. Then we greedily select the trajectory $[\mathbf{x_g}(t),\mathbf{u_g}(t)]$ such that
\begin{equation}\label{eq:cost_prediction}
\argmin_{g}{\mathbf{\hat{x}}^\top Q(t)\mathbf{\hat{x}}+\mathbf{\hat{u}}^\top R(t)\mathbf{\hat{u}}}.
\end{equation}

\section{EXPERIMENTS}
In our experiments we employ a simple two spherical fingers parallel gripper shown in \fig{fig:gripper}. We demonstrated to the system a pinch grasp over rectangular shaped object as shown in \fig{fig:contact_model}. The learnt contact model (\sect{sec:contact_model}) computes a pdf over the poses for links $L1$ and $L2$ given the object's surface features.  At prediction time, the query density described in \sect{sec:query} will attempt to sample candidate grasps which have similar characteristics to the learnt contact model. 

\subsection{Experimental setup}\label{sec:setup}

\begin{figure}[!t]
\centerline{
\includegraphics[width=.495 \textwidth]{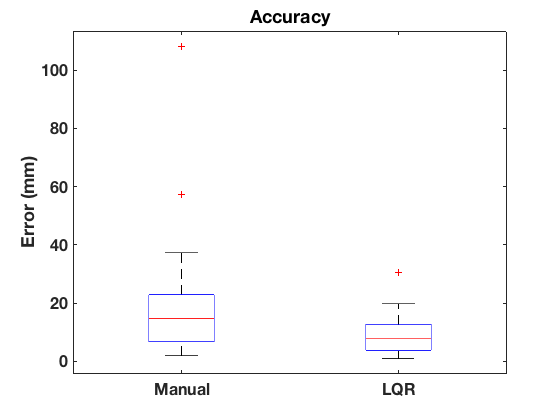}
}
\caption{Position error with respect to the target grasp in mm. Left: the performance of the participants attempting to reach the desired grasp manually. Right: the performance with the help of the LTV-LQR controller. Red lines show the mean error, the bars the standard deviation and the red crosses outliers.  
}
\label{fig:accuracy}
\end{figure}

We collected data from 4 participants: two males and two females between 20 and 40 years of age. All the participants have no history of neuromuscular disorders. Each subject was presented with a random generated scene, as in \fig{fig:grasps}, and a random object (i.e. rectangular shape) was selected. We asked the subjects to move the gripper as to grasp the selected object by using the numerical pad of a standard keyboard. The keys 2, 4, 6, 8, were used for down, left, right and up movements respectively and 1, 3, 7 and 9 for the diagonal movements.

\begin{figure}[!t]
\centerline{
\includegraphics[width=.495 \textwidth]{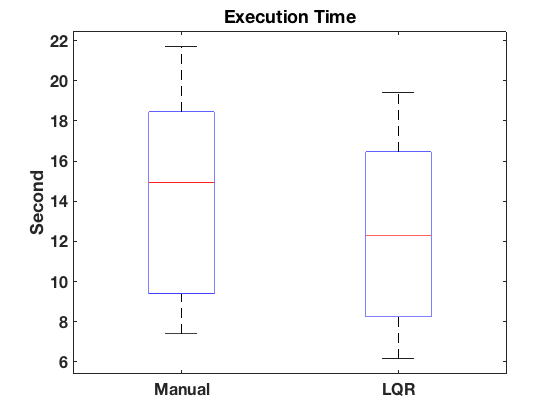}
}
\caption{Execution time for all the participants in seconds. Red lines show the mean error, the bars the standard deviation and the red crosses outliers.  
}
\label{fig:time}
\end{figure}

We collected a total of 80 trials. Each subject performed 20 trials: 10 trials controlling the gripper manually and 10 trails using the LTV-LQR. We analysed 2 metrics: 1) accuracy in achieving the grasp and 2) execution time form the first control input until the grasp was achieved. Figure~\ref{fig:accuracy} shows that our approach is superior to the manual control and allows the user to reach with better accuracy the desired target. In \fig{fig:time}, we show the execution time for all the trials. Our controller allows the user to drive the end-effector towards the target faster, this is because imprecise inputs from the users are filtered out by the controller which still generate a smooth trajectory. However, some reduced performance in time has been registered when the system, especially to the earlier stages of the movement, selects the wrong candidate trajectory. Nonetheless the user is capable to generate a sequence of inputs, i.e. driving the end-effector towards a different candidate, to inform the system about his/her intention which yields to a change in the selected trajectory to support the user.

Although the simplistic scenario, the results encourage the development of AI controllers for semi-autonomous robots. Context-awareness allows the system to understand the scene and generate candidate solutions even in novel scenarios. At the same time, user-awareness interprets the input signals of the user to identify the task, i.e. which object to grasp. 

The benefits of our approach are twofold: 1) reduce the complexity of controlling the device, thanks to the ability of the system to infer orientation and fingers' configurations, and 2) imprecise inputs will be filtered by the controller to obtain a smooth trajectory.

\section{CONCLUSION AND FUTURE WORK} 
\label{sec:conclusion}

The presented approach is a step forward to improve accuracy and usability of semi-autonomous robots such as human operated manipulator for nuclear waste disposal. The key idea is to develop a system that is simultaneously context- and user-aware. We implement the context-awareness by enabling the system to generate a set of candidate grasps on novel scenes, and by planning nominal reach-to-grasp trajectories. For each trajectory a linear time-variant LQR controller is created that identifies the user's intention from motion commands and guide the user to the intended grasp. The user's  burden is limited to control the end-effector in 2D while the orientation and the pose of the fingers are inferred by the system according to the selected grasp. The system also is capable to recover from wrong decisions when the selected target grasp is not the same as the one chosen by the user.    

The results have demonstrated that in our simple scenario the user is capable to reach a desired target faster and with better accuracy than on manual control. Future work will be addressed to extend this type of control to a 3D scenario in which a user controls a real robot arm for tasks like pick and place, and to investigate whether the user's cognitive burden is reduced by our approach. Additionally, we are also interested to investigate different types of input control, such as EMG signals, to extend this work to benefit patients with neuromuscular disorders and amputees. 

%

\addtolength{\textheight}{-12cm}   


\bibliographystyle{plain}
\bibliography{IROS2018}

\end{document}